# HUMAN MOOD DETECTION FOR HUMAN COMPUTER INTERACTION

Ms. Priti B. Badar, Ms. Urmila Shrawankar
*Dept. of Computer Science. and Engineering. (MTech. Final year)*
*G. H. Raisoni College Engineering, Nagpur*
*Nagpur, Maharashtra, 440032, India*
*Priti_badar@yahoo.co.in*

## ABSTRACT

*In this paper we propose an easiest approach for facial expression recognition. Here we are using concept of SVM for Expression Classification. Main problem is sub divided in three main modules. First one is Face detection in which we are using skin filter and Face segmentation.*

*We are given more stress on feature Extraction. This method is effective enough for application where fast execution is required. Second, Facial Feature Extraction which is essential part for expression recognition. In this module we used Edge Projection Analysis. Finally extracted feature's vector is passed towards SVM classifier for Expression Recognition. We are considering six basic Expressions (Anger, Fear, Disgust, Joy, Sadness, and Surprise)*

## 1. Introduction

The computers are now able to read a text, to recognize words, sentences, but they are not yet able to understand! The recognition of facial expressions could be very useful in the dialogue between human and machine. It could be a way for the machine to get information on what you think or how you feel and so being able to answer to you in the right way. Even if we often do not realize it, facial expressions are a very important way to communicate with somebody else.

At this moment there are many researchers involved in projects concerning non-verbal communication. For Face detection Hausdorff distance is used as a similarity measure between a general face model and possible instance of the object within the image [10]. Next one is the system consists of two level, fist one is component classifier second is matches it with face model [11]. In another method the concept of Active Contour Models (ACMs) and Genetic Algorithm are used [12].

Different Classifiers are in used to get accuracy to solve the problem of Facial Expression Recognition. Euclidean classifier. This classifier assigns an object to the class with the closest mean (computed with Euclidean distance). Recognition accuracy: 80% recognition.

The nearest neighbor classifier, the object goes in the class which proposes the nearest point to this object. Recognition accuracy: relatively little degradation in recognition under partial face occlusion or tracker noise.

Hidden Markov models. It works on basis of state transition. Recognition accuracy: 85% recognition.

The neural networks, for example the multi-layer perceptron, are a supervised learning algorithm. The training of this multi-layer perceptron consists of computing the weights of the connections between the different cells. Recognition accuracy: 93.3% recognition

Rule-based expert system. Recognition accuracy: 91% recognition

When tackling the problem of Facial Expression recognition system, one generally decomposes it in three sub problems:
• Face Detection
• Facial Features Detection
• Expression Recognition/Classification

The first thing to do when one wants to design a facial expression recognition system is to select the Face under which experiments will have to be run. In our case, as we want the system to be fully automatic, we have to start by detecting the user's face inside the scene. Camera captures the image, which includes the human face. Then human face will be detects from image. Features, which are required for facial expression recognition, will be detected and then next module of the system detects the Expression.

## 2. Finding Face from Image

First we have to find the Face from the image. Although it seems like an easy problem at first glance, we quickly realized that the high variability in the types of faces encountered makes the automatic detection of the face a tricky problem.

Two steps are there to detect the face
1. Detect the region which likely to contain the human skin in the color image.
2. Then extract information from these regions, which might indicate the location of face in the image. Now the skin detection is performed using a skin filter, which relies on color and texture information.

### 2.1 Skin Filter:-

Skin filter uses hue and saturation as the main arguments. Using Constraints on hue and saturation region of skin can be marked.

## 2.2 Proposed Face Detection Method

To overcome the problems being faced in most of the earlier methods:
- At first, completely unimportant colors are eliminated from image (those which can't represent a face) as shown in Fig 1. All insignificant colors are replaced with white color.
- Image is then converted into grayscale picture.
- It is filtered with a median filter.

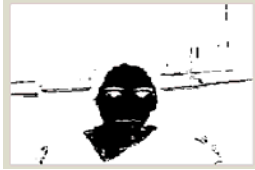

Figure 1. Grayscale image

- Using Skin filter Skin region are Segmented as shown in Fig 2.

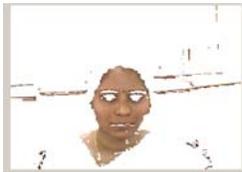

Figure 2. Skin region segmentation

- Edges are traced in the image with significance grays as shown in Fig 3.
- Within preserved regions the algorithm searches for circles using Hough transform.
- For each region the best possible circle is found.

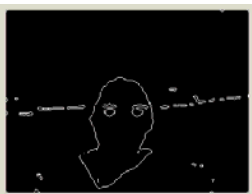

Figure 3. Edges is trace from Image

- For face candidate confirmation, color information of the whole face is used. With the help of this information the algorithm also (loosely) predicts the probability of a face.
- Finally, using some heuristic rules, the algorithm tries to eliminate falsely classified faces.

## 3. Feature Extraction

Now we require Features which are essential for expression. Following Feature plays important role to identify the human expressions.
- Eyebrows
- Lips
- Nose

We are using the Edge projection analysis for features extraction. Using integral projections of the edge map of the face image facial features can be extracted.

Let $I(x, y)$ be the input image. Vertical and horizontal projection vectors in the rectangle $[x1, x2] \times [y1, y2]$ are defined as in Fig. 4

A typical human face follows a set of anthropometric standards, which have been utilized to narrow the search of a particular facial feature to smaller regions of the face.

We use the following generic steps for the facial feature extraction from the localized face image—

1. An approximate bounding box for the feature is obtained using the anthropometric standards.
2. Sobel edge map (fig 4.1) is computed to obtain edges along the boundary of the feature.
3. The integral projections $V(x)$ and $H(y)$ are calculated on the edge map (fig 4.2, 4.3, and 4.4).
4. Median filtering followed by Gaussian smoothing smooths the projection vectors so obtained. Higher value of projection vector at a particular point indicates higher probability of occurrence of the feature. The relative probability $E(i)$ of the ith region containing the feature is calculated as

$$E(i) = \sum_{y=yi}^{y=yi+1} H(y) \, w(y)$$

5. The bounding box so obtained is processed further to get an exact binary mask of the feature.

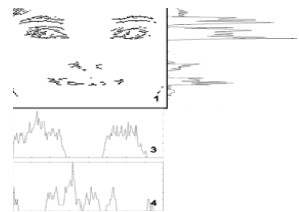

Figure 4. Generic steps to get bounding box. (1) Edge map, (2) Vector H (y), (3) Vector V (x) in upper half portion,(4) Vector V (x) in lower half portion,

### 3.1. Eyebrow

- The approximate bounding box is the top half of the face. The generic steps uses horizontal sobel edges to compute bounding box containing *eye* and *eyebrow*.
- The segmentation algorithm cannot give bounding box for the eyebrow exclusively because the edges due to eye also appear in the chosen bounding box.

- Brunelli suggests use of template matching for extracting the eye.
- Eyebrow is segmented from eye using the fact that the eye occurs below eyebrow and its edges form closed contours obtained by applying *Laplacian of Gaussian* operator at zero threshold.
- These contours are filled and the resulting image containing masks of eyebrow and eye is morphologically filtered by horizontally stretched elliptic structuring elements.
- From the two largest filled regions, the region with higher centroid is chosen to be the mask of eyebrow.

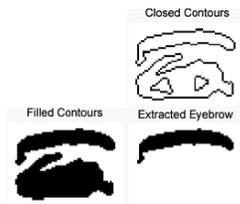

Figure 5. Post processing of approximate bounding box to get exact binary mask of eyebrow.

### 3.2. Lip

- The generic steps calculate edge maps on the image.
- Edges for lips occur both in horizontal and vertical direction.

### 3.3. Nose

- The approximate bounding box for the nose lies between the eyes and the mouth.
- The generic steps uses vertical sobel edges to compute the vertical position, which is required as a reference point on face.

## 4. Classifier

Classifier takes Feature vector as a input and classify them to one of the six basic expressions.

### 4.1 Feature Vector and Classification

A spatio-temporal representation of the face is created, based on geometrical relationships between features using Euclidean distance.
Seven parameters form the feature vector F—
F = {He, We, Hm, Wm, Rul, Rll, NL}
He: Height of eyebrow
We: brows distance
Hm: mouth height
Wm: mouth width
Rul: upper lip curvature
Rll: lower lip curvature

### 4.2 What is Support Vector Machine?

Support vector machines (SVMs) are a set of related supervised learning methods used for classification and regression. They belong to a family of generalized linear classification. A special property of SVMs is that they simultaneously minimize the classification error and maximize the geometric margin. Hence they are also known as maximum margin classifiers. SVM is particularly a good tool to classify a set of points which belong to two or more classes. SVMs are based on statistical learning theory and try to find the biggest margin to separate different classes.

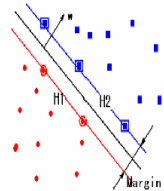

Figure 6. Support Vector and Margin

SVM embed data into a high dimensional feature space. This method uses the hyper plane that separates the largest possible fraction of points of the same class on the same side, while it maximizes the distance of either class from the hyper-plane.

### 4.3 Use of Support Vector Machine in our System

Machine learning algorithms receive input data during a training phase. Build a model of the input and output a hypothesis function that can be used to predict future data. SVM algorithms separate the training data in feature space by a hyper plane defined by the type of kernel function used. They find the hyper plane of maximal margin, defined as the sum of the distances of the hyper plane from the nearest data point of each of the two classes. The margin represents a measure of class separation efficiency and is defined as the Euclidean distance between the data and the separating hyper plane. The size of the margin bounds the complexity of the hyper plane function and hence determines its generalization performance on unseen data.

## 5. Conclusion

In this paper we presented a Real-time system for Facial Expression Analysis. To implement this technique we used MATLAB's Image Processing Toolbox. For testing this, we used our own data based which gives 75% accuracy in First module. And the Result of Classification is shown in Table 1.

Table 1: Recognition accuracy of SVM classification

| Expression | Percentage |
|---|---|
| **Anger** | 65.5% |
| **Disgust** | 64.3% |
| **Fear** | 66.7% |
| **Joy** | 91.7% |
| **Sadness** | 62.5% |
| **Surprise** | 83.3% |
| **Average** | 71.5% |

- To detect Face and Extract Facial Features we used Simple, fast and low cost method. The two basic limitations of this method are Input image must have high enough resolution. The face must be big enough. It is sensitive to the complexion. But this limitation does not affect more on overall performance of our system. This system may be use in academic and applied fields including
  1. Clinical psychology and psychiatry
  2. Child development,
  3. Political science and Biomedical Applications
  4. Computer systems
  5. Speech recognition
  6. Security systems and Lie detection..
  7. Video compression
  8. Facial animation

Most of the educational areas are trying to build most powerful and intelligent tutorial system. By using human mood detection system we can make more powerful tutorial. In our future work we will concentrate on intelligent tutorial system using Facial Expression Recognition System.

## 6. References

[1] P. Viola and M. J. Jones, "Robust Real-Time Face Detection," *International Journal of Computer Vision*, vol. 57, pp. 137-154, 2004.

[2] C. J. C. Burges, "A Tutorial on Support Vector Machines for Pattern Recognition," in *DataMining and Knowledge Discovery*, vol. 2. Boston: Kluwer Academic Publisher, 1998, pp. 121-167.

[3] C. C. Chang and C. J. Lin, "LIBSVM: a library for support vector machines," 2001.

[4] Chao Fan, Hossein Sarrafzadeh, Farhad Dadgostar1, and Hamid olamhosseini, "Facial Expression Analysis by Support Vector Regression", in 2Department of Electrical & Electronic Engineering
Auckland University of Technology
Private Bag 92006, Auckland 1020

[5] B. Fasel, et al. Automatic Facial Expression Analysis: A Survey. Pattern Recognition, 36, 259-275, 2003

[6] F. Bourel, et al. Robust Facial Expression Recognition Using
a State-Based Model of Spatially-Localised Facial Dynamics. Proc Fifth IEEE Int'l Conf on Automatic Face and Gesture Recognition (FGR-02), Washington D.C., 113-118, 2002.

[8] Ashutosh Saxena, Ankit Anand, Prof. Amitabha Mukerjee, "ROBUST FACIAL EXPRESSION RECOGNITION USING SPATIALLY LOCALIZED
GEOMETRIC MODEL", in International Conference on Systemics, Cybernetics and Informatics, February 12–15, 2004

[9] Kanade T., et al. Comprehensive Database for Facial Expression Analysis. Proc. Fourth IEEE Int'l Conf on Automatic Face and Gesture Recognition (FG'00). Grenoble, France. March 2000.

[10] Oliver Jesorsky, Klaus J. Kirchberg "Robust Face Detection using the Hausdorff Distance", in University of Liubljana, Slovenia.

[11] Thomas serre, Berend Heisele, "Component – based Face Detection", in University of Siena, Italy.

[12] Ambellouis Sebastien & Megherbi Najla "A Genetic Snake Model to Automatic Human Face and Head Boundary Detection" Laboratories Electronique, ondes et Signaux pour les Transports INRETS , FRANCE.